\documentclass[11pt]{article}

\usepackage[preprint]{acl}

\usepackage{times}
\usepackage{latexsym}
\usepackage{amsmath}
\usepackage{amssymb}
\usepackage{amsfonts}
\usepackage{graphicx}
\usepackage{algorithm}
\usepackage{algpseudocode}
\usepackage{tabularx}
\usepackage{booktabs}
\usepackage{array}
\usepackage{float}

\usepackage[T1]{fontenc}
 
\usepackage[utf8]{inputenc}

\usepackage{microtype}

\usepackage{inconsolata}

\usepackage{graphicx}
\usepackage{float}
\usepackage{caption}
\usepackage{subcaption}  %
\usepackage{xspace}

\renewcommand{\tt}{\textsc{TimpaTeks}\xspace}
\newcommand{\ts}{\textsc{TimpaSteer}\xspace}
\newcommand{\imdb}{{\texttt{IMDB}}\xspace}
\newcommand{\catdog}{{\texttt{CatDog}}\xspace}
\newcommand{\elifive}{{\texttt{ELI5}}\xspace}

\title{TimpaTeks: Token Localization for Masked Diffusion Language Model Text Editing}

\author{
 \textbf{Ryandito Diandaru},
 \textbf{Ahmed Elshabrawy},
 \textbf{Fadli Aulawi Al Ghiffari},\\
 \textbf{Ikhlasul Akmal Hanif},
 \textbf{Alham Fikri Aji}
\\
 MBZUAI
\\
 {\texttt{\small\{Ryandito.Diandaru, Ahmed.Elshabrawy, Fadli.Ghiffari, Ikhlasul.Hanif, Alham.Fikri\}@mbzuai.ac.ae}}
}

\begin{document}
\maketitle
\begin{abstract}
We study in-place text modification with diffusion language models, where selected tokens in a source text are masked and regenerated. Our method is \tt, a training-free method that exploits the change in token likelihoods under base and target prompts to identify tokens that should be revised to match the desired outcome. Across sentiment transfer on \imdb, synthetic concept transfer on \catdog, and a harder, less cleanly-cut attribute transformation on \elifive, \tt provides a controllable trade-off between target success and source preservation. DLM instruction and autoregressive baselines reliably achieves the target transformation, whereas targeted editing with does so with only the minimal, necessary edits. Targeted masking also outperforms random remasking on \elifive. An activation steering-based baseline, \ts, frequently under-edits, suggesting that remasking detection using token probabilities is more reliable. We release our code.
\end{abstract}

\section{Introduction}
\begin{figure*}[t]
    \centering
    \includegraphics[width=0.9\textwidth]{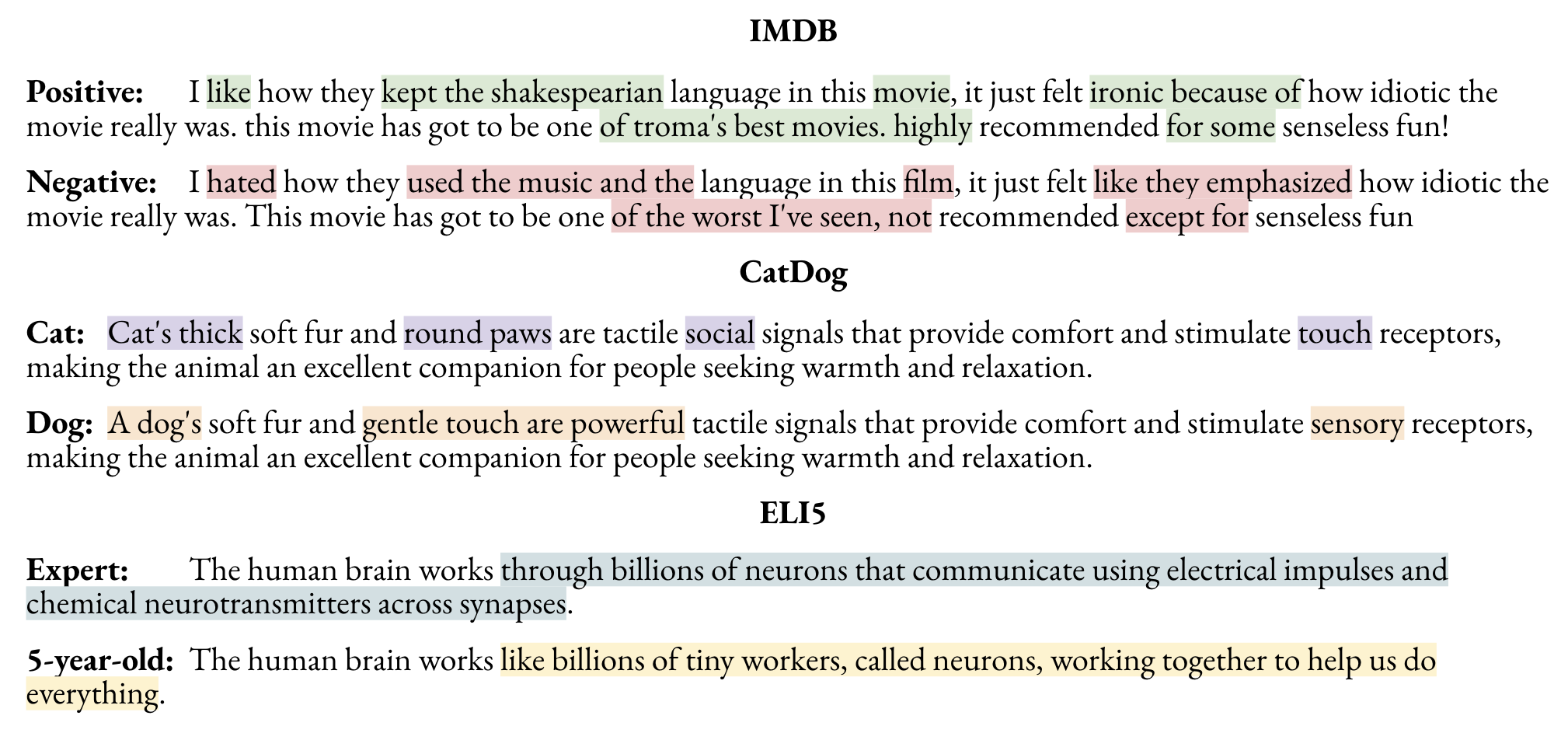}
    \caption{Example run of the \tt~method.}
    \label{fig:timpa_illust}
\end{figure*}
Language models can modify text by retaining selected input tokens and generating replacements or insertions \cite{malmi-etal-2019-encode, mallinson-etal-2020-felix}. Text modification is also commonly performed by prompting an instruction-tuned model to rewrite the input according to a desired attribute \cite{raheja-etal-2023-coedit}. Alternatively, activation steering guides the output without retraining by intervening in the model's intermediate activations at inference time \cite{rimsky-etal-2024-steering}.

On the other hand, the parallel generation mechanism of Diffusion Language Models (DLMs) unlocks many opportunities to exploit them for this use case. Unlike autoregressive (AR) models that generate text from left to right, DLMs iteratively denoise masked sequences using bidirectional context. This allows them to predict tokens in a flexible order rather than one at a time \cite{sahoo2024simpleeffectivemaskeddiffusion, nie2025largelanguagediffusionmodels}, making them a natural fit for this setting: by masking only selected tokens, they can then regenerate those positions using the unchanged surrounding text as context. These properties make DLMs a promising setting for \textit{in-place} text modification. However, it remains underexplored, so we ask: \textit{Can DLMs be used to modify a sentence in-place such that it manifests a different concept while retaining structure and coherence?}

To answer this, we introduce \tt, a token localization strategy that exploits changes in token likelihoods. \tt first identifies tokens that \textit{do not} fit the target transformation and remasks them. After remasking, the partially masked sequence produced undergoes the DLM denoising process. Additionally, since many works studying sequence modification in AR models use activation steering \cite{rodriguez2025controlling, lee2025programming, templeton2024scaling}, we also investigate how activation steering would work in the formulated \textit{in-place} text editing task with DLMs, however we demonstrate that it is less successful than \tt.

As seen in Figure~\ref{fig:timpa_illust}, we find \tt~to be particularly effective. We compare \tt against autoregressive editing and activation steering baselines and find that \tt balances transformation success and source sentence attributes preservation in sentiment (\imdb), concept (\catdog), and audience (\elifive) datasets. Our contributions are, hence, the following:
\begin{itemize}
    \item We introduce \tt, a novel framework for automatically modifying text \textit{in place} with DLMs through token-level probabilities and evaluate it across three datasets and multiple metrics using statistical significance tests.
    \item We provide a comparison of \tt against DLM instruction, autoregressive baselines, and activation steering and analyze their tradeoffs.
\end{itemize}

\section{Related Work}

Prior work has framed text modification as editing an existing sequence rather than generating a new one from scratch. LaserTagger predicts a keep or delete operation for each input token, along with the phrases to insert \cite{malmi-etal-2019-encode}. FELIX extends this approach by selecting and reordering the retained tokens and then infilling the missing content with a masked language model \cite{mallinson-etal-2020-felix}. CoEdIT instead instruction-tunes a language model to perform a range of rewrites, including simplification, style transfer, and grammatical error correction, given natural-language editing instructions \cite{raheja-etal-2023-coedit}.

Activation steering is also one approach to modifying autoregressive language-model outputs without fine-tuning. Motivated by evidence that semantic attributes can correspond to linear directions in representation space \cite{mikolov-etal-2013-linguistic,park2023the}, early methods constructed directions from contrastive activations and added them to the residual stream during generation \cite{turner2024steeringlanguagemodelsactivation,rimsky-etal-2024-steering}. Activation-based control has also been extended to masked DLMs. \citet{shnaidman2026activationsteeringmaskeddiffusion} apply projection removal across diffusion timesteps and analyze the effects of intervention layers, timesteps, and token subsets.

More recently, FineEdit \cite{zeng-etal-2025-bridging} and HyperEdit \cite{zeng-etal-2026-hyperedit} adapt autoregressive language models to instruction-based editing through supervised fine-tuning. Diffusion models have also been applied to text editing: EdiText performs controllable editing with a latent diffusion model trained on attribute-labeled data \cite{lee-etal-2025-editext}, and Diff4TST fine-tunes a masked diffusion language model for text style transfer \cite{ma-etal-2026-diff4tst}. In both cases, editing toward a new target attribute requires additional training. To the best of our knowledge, \tt~ is the first method that performs text editing with a pretrained masked diffusion language model that requires no parameter updates, including when the target attribute was not seen during training.

\section{Method}\label{methodology}

\paragraph{Problem Formulation}

Given a source sequence $\mathbf{x}=(x_1,\ldots,x_N)$ and a target instruction $q_{\mathrm{target}}$, our goal is to modify $\mathbf{x}$ toward the requested attribute while preserving all positions that do not require modification. The full text editing consists of two stages. The first step is \tt~: an autoregressive scorer identifies words that become less likely under the target instruction than under a neutral instruction. These words are sampled for replacement. Second, a target-conditioned DLM regenerates only the selected positions using the unchanged source tokens as bidirectional context.

We intentionally do not use DLMs themselves as a scorer model and opt to use a separate autoregressive model due efficiency reasons. To get the likelihoods, each token would need to be forwarded through the model conditioned on the rest of the tokens in the sequence. This would need to be done separately for each token. This overhead in the identification process if done using DLMs would outweigh the benefits of editing only the necessary tokens.

\begin{figure*}[t]
    \centering
    \includegraphics[width=0.9\textwidth]{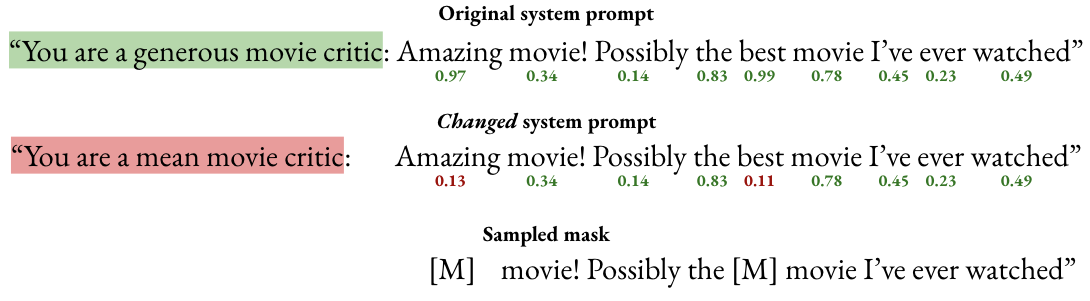}
    \caption{\tt selects tokens by identifying which tokens become incompatible when the system prompt is changed. In this illlustration, \texttt{"amazing"} and \texttt{"best"} decreased in likelihood because they no longer fit \texttt{"a mean critic"} and thus are more likely to be remasked}
    \label{fig:likelihood_illust}
\end{figure*}

\paragraph{Target-Incompatibility Scoring}\label{detect}
Let $q_{\mathrm{base}}$ and $q_{\mathrm{target}}$ denote neutral and target-conditioned prompts, respectively. For each source token $x_i$, an autoregressive scorer computes
\begin{align*}
P_{\mathrm{base}}[i]
&=
p_{\mathrm{AR}}\!\left(
x_i \mid q_{\mathrm{base}},\mathbf{x}_{<i}
\right),\\
P_{\mathrm{target}}[i]
&=
p_{\mathrm{AR}}\!\left(
x_i \mid q_{\mathrm{target}},\mathbf{x}_{<i}
\right).
\end{align*}
We define the target-conditioned log-likelihood change as
\[
\Delta_i
=
\log P_{\mathrm{target}}[i]
-
\log P_{\mathrm{base}}[i].
\]
A negative value indicates that the observed token becomes less compatible with the context under the target instruction, making it a candidate for replacement. The intuition behind why this should reliably identify tokens that need to be remasked is in Figure~\ref{fig:likelihood_illust}

\paragraph{Probabilistic Word Selection}

Because a word may contain multiple tokens, we aggregate token-level changes into a word-level log-likelihood ratio. For each word $w$, let $\mathcal{I}_w$ denote its autoregressive token indices:
\[
\Delta_w
=
\sum_{i\in\mathcal{I}_w}\Delta_i.
\]
The sum corresponds to the change in the word's joint log-likelihood under the target condition.

We convert this score into a masking probability:
\[
p_w
=
\tanh\!\left(
\frac{\max(0,-\Delta_w-m)}{\tau}
\right),
\]
where $m\geq0$ is a margin and $\tau>0$ controls the sharpness of the distribution. Only words whose target-conditioned likelihood decreases by more than $m$ receive nonzero masking probability. We then sample
\[
z_w\sim\operatorname{Bernoulli}(p_w).
\]
If $z_w=1$, all DLM tokens corresponding to $w$ are selected for regeneration.

\paragraph{Masking controls.}
We compare stochastic TimpaTeks selection with two controls. \textsc{Deterministic} masks every word satisfying $\Delta_w<-m$, testing whether stochastic selection is necessary. \textsc{Random} uniformly selects a matched number of words, testing whether performance depends on selecting target-relevant positions rather than merely regenerating the same amount of text.

\paragraph{In-place Diffusion Regeneration}

Let $R$ denote the DLM-token positions corresponding to the sampled words. We replace only these positions with the model's mask token, producing $\mathbf{x}^{\mathrm{mask}}$. All positions outside $R$ remain fixed throughout generation.

The DLM then iteratively fills the masked positions while conditioned on $q_{\mathrm{target}}$. At step $r$, a refill schedule selects $k_r$ masked positions, and the model commits the corresponding token predictions. This continues until all selected positions are filled or the budget of $u$ denoising steps is exhausted. The procedure adapts the sampling process of LLaDA \cite{nie2025largelanguagediffusionmodels}, but applies it directly to selected source positions rather than generating a separate output sequence.

\begin{algorithm}[t]
\caption{\tt}
\label{alg:timpateks}
\begin{algorithmic}[1]
\Require Source $\mathbf{x}$; prompts $(q_{\mathrm{base}},q_{\mathrm{target}})$; AR scorer $g$; DLM $f$; margin $m$; temperature $\tau$; refill steps $u$
\Ensure Modified sequence $\tilde{\mathbf{x}}$

\State $\mathbf{P}_{\mathrm{base}}
    \gets \textsc{ScoreAR}(g,\mathbf{x},q_{\mathrm{base}})$
\State $\mathbf{P}_{\mathrm{target}}
    \gets \textsc{ScoreAR}(g,\mathbf{x},q_{\mathrm{target}})$
\State $\boldsymbol{\Delta}
    \gets \log\mathbf{P}_{\mathrm{target}}
    - \log\mathbf{P}_{\mathrm{base}}$
\State $R \gets
    \textsc{SampleWords}(\boldsymbol{\Delta},\tau,m)$
\State $\tilde{\mathbf{x}}
    \gets \textsc{Mask}(\mathbf{x},R)$
\State $(k_1,\ldots,k_u)
    \gets \textsc{RefillSchedule}(\tilde{\mathbf{x}},u)$

\For{$r=1$ to $u$}
    \If{$\textsc{NoMasks}(\tilde{\mathbf{x}})$}
        \State \textbf{break}
    \EndIf
    \State $\mathbf{y}
        \gets f(\tilde{\mathbf{x}};q_{\mathrm{target}})$
    \State $P
        \gets \textsc{SelectUnmask}
        (\tilde{\mathbf{x}},\mathbf{y},k_r)$
    \State $\tilde{\mathbf{x}}
        \gets \textsc{FillMasks}
        (\tilde{\mathbf{x}},\mathbf{y},P)$
\EndFor

\State \Return $\tilde{\mathbf{x}}$
\end{algorithmic}
\end{algorithm}

\paragraph{Activation-based Comparison: \ts}
\label{sec:timpa-steer}

We additionally test whether activation-space signals can replace the likelihood-based localization and prompted regeneration used by \tt. This comparison, which we call \textsc{TimpaSteer}, constructs layer-wise steering directions from the difference between mean representations of contrasting concepts.

For token localization, \textsc{TimpaSteer} measures the cosine similarity between each token representation and the corresponding steering directions. Tokens negatively aligned with the target direction receive higher masking probabilities. During regeneration, the same directions are added to the DLM residual stream:
\[
\tilde{h}_i^\ell(\mathbf{x}_t)
=
h_i^\ell(\mathbf{x}_t)+\alpha S^\ell.
\]
This replaces both components of \tt: activation similarity replaces conditional likelihood scoring, and activation intervention replaces target-prompted denoising. We therefore treat \textsc{TimpaSteer} as an end-to-end activation-based comparison rather than a second proposed method. Full implementation details are provided in Appendix~\ref{sec:timpa-steer-details}.

\section{Experimental Setup}
\label{sec:experimental-setup}

We evaluate three questions: (1) What trade-offs are there between full-sequence rewriting and \textsc{Timpa} methods? (2) Does sampled masking identify more useful edit positions than random or deterministic masking? (3) Can \ts provide a comparable alternative?

\subsection{Tasks and Datasets}

We evaluate on three text-transformation tasks. \imdb~\cite{imdb2011} contains movie reviews labeled for positive or negative sentiment. We use it to evaluate bidirectional sentiment transfer. \catdog~is a synthetic dataset for categorical concept transfer between cats and dogs (details in Appendix~\ref{sec:catdog-details}). \elifive~contains technical explanations targeting three ordered audience levels: a five-year-old child, a high-school student, and a subject-matter expert with a PhD. While \imdb and \catdog primarily require changing a salient attribute, \elifive requires finer-grained control over terminology, abstraction, and level of detail. We would like to highlight that \elifive serves as a more complex benchmark than \imdb and \catdog in the sense that it does not involve \textit{cleanly cut} concepts within it, thus making it a robustness test at the same time.

\subsection{Evaluation Metrics}

Our evaluation distinguishes \emph{target control}, which measures whether the requested transformation is achieved, from \emph{source preservation}, which measures whether unrelated content and structure remain intact.

\paragraph{\imdb and \catdog.}
We evaluate three metrics. \textbf{Target success} measures whether the output exhibits the requested sentiment or concept using a local classifier based on label-token probabilities. \textbf{Faithfulness} measures whether the substantive source content is preserved. \textbf{Structure preservation} evaluates whether the organization and grammatical structure of the source remain intact.

\paragraph{\elifive.}
We evaluate \textbf{factuality}, \textbf{faithfulness}, and \textbf{structure preservation}. We additionally measure whether the outputs recover the intended ordering of audience complexity.

For source instance $x_i$, let $y_i^{(a)}$ denote its rewrite for audience level $a\in\{1,2,3\}$, corresponding to five-year-old, high-school, and PhD-level audiences. The three outputs are randomly permuted before being presented to the judge, which ranks them from simplest to most advanced. Let $\hat{r}_{i,a}$ be the assigned rank and $r_a^\star=a$ the intended rank. We compute
\[
\operatorname{OrderMAE}_i
=
\frac{1}{3}
\sum_{a=1}^{3}
\left|\hat{r}_{i,a}-r_a^\star\right|.
\]
Lower values indicate better recovery of the intended progression. This metric evaluates relative audience separation rather than absolute suitability for each population.

We calculate two versions of this metric. \textbf{LLM Order} uses rankings assigned by an LLM judge, while \textbf{Flesch Order} ranks outputs using Flesch Reading Ease. Factuality, faithfulness, structure preservation, and LLM Order are evaluated using an LLM judge. The complete prompts and scoring rubrics are provided in Appendix~\ref{sec:prompts-used}.

\subsection{Compared Methods}

\tt~ is our proposed method. It uses target-conditioned likelihood changes to identify incompatible source words, samples words for replacement, and regenerates only the selected positions through target-prompted diffusion denoising. We evaluate three selection policies:

\begin{itemize}
\itemsep0em
    \item \textbf{\tt} uses the stochastic masking configuration selected on held-out validation data.
    \item \textbf{\tt--Deterministic} masks every word whose likelihood change crosses the target-incompatibility threshold. This tests whether stochastic selection is beneficial.
    \item \textbf{Random Mask} replaces 50\% of source words uniformly at random and regenerates them using the same target-prompted DLM. This tests whether performance depends on selecting target-relevant positions rather than merely regenerating part of the source.
\end{itemize}

For our main experiments, due to our limited compute budget, we use only one paraphrase for the source and target prompts. However, we have run a small experiment to prove that these prompts are robust towards paraphrasing, We ran \elifive across 5 paraphrases (100 examples; fixed seed) and found low standard deviations: factuality, faithfulness, structure preservation, and Flesch MAE were 0.033, 0.054, 0.075, and 0.051, respectively.

\paragraph{Full-rewrite baseline:}
\textbf{Instruction} refers to using the DLM backbone itself to rewrite the complete source according to the target instruction. We also run two autoregressive models: \textbf{GPT-6 Astra} \cite{openai2026gpt6astra} (\textit{state of the art} baseline) and \textbf{Qwen3.5 9B} \cite{qwen35} (similarly sized autoregressive model) as additional reference points. Unlike \tt, they generate a separate output sequence and places no hard constraint on retaining individual source tokens. Thus, it represents the naive way to perform text modification with instruction prompting in DLMs. It provides a strong reference for target control, while \tt evaluates whether restricting generation can improve preservation.

\paragraph{Activation-based comparison.}
\textbf{\ts} replaces likelihood-based masking and target-conditioned regeneration with activation-space token scores and activation intervention. We evaluate validation-selected, deterministic, and random-mask configurations. \ts is included as an alternative end-to-end design ablation rather than a second proposed method.

This produces seven experimental conditions: Instruction; \tt--Selected; \tt--Deterministic; \tt--Random; \ts--Selected; \ts--Deterministic; \ts--Random;

\subsection{Models and Implementation}

We use \texttt{LLaDA-8B-Instruct}~\cite{nie2025largelanguagediffusionmodels} as the masked DLM for all methods. For \tt, \texttt{Qwen2.5-32B-Instruct}~\cite{qwen2025qwen25technicalreport} provides the autoregressive token-likelihood scores. The same model provides the label-probability classifier used for target-success evaluation. We use \texttt{gpt-4.1-mini}~\cite{openai2025gpt41} as the LLM judge for factuality, faithfulness, structure preservation, and audience ordering. We run every stochastic condition with seeds 42, 43, and 44.

\subsection{Validation and Hyperparameter Selection}

All hyperparameters are selected using a held-out validation set of 10 instances. The test set is not used during model or hyperparameter selection.

\paragraph{\textbf{\tt}}
We search over masking temperatures
$\{0.25,0.5,1.0,2.0\}$ and margins
$\{0.001,0.1,0.25,0.5\}$, producing 16 configurations. The validation search selects $\tau=0.5$ and $m=0.1$, which we use as the default \tt configuration. On the validation data, this configuration obtains factuality, faithfulness, and structure-preservation scores of $4.90$, $4.07$, and $4.37$, respectively, while tying for the lowest Order MAE at $0.47$.

\paragraph{\textbf{\ts}}
Because TimpaSteer introduces additional parameters for activation strength, intervention layers, token detection, and refill behavior, we tune it separately for each dataset--vector-basis pair using a 16-round greedy coordinate search. Candidates are selected using a weighted combination of target control and preservation metrics. The search evaluates 250 trials across five dataset--vector-basis tracks, after which all configurations are fixed before test evaluation. Final runs use label-derived vectors and a 32-step refill budget. The complete search space, selection weights, and procedure are reported in Appendix~\ref{sec:timpa-steer-search}.

\begin{figure*}[t]
    \centering
    \includegraphics[width=\textwidth]{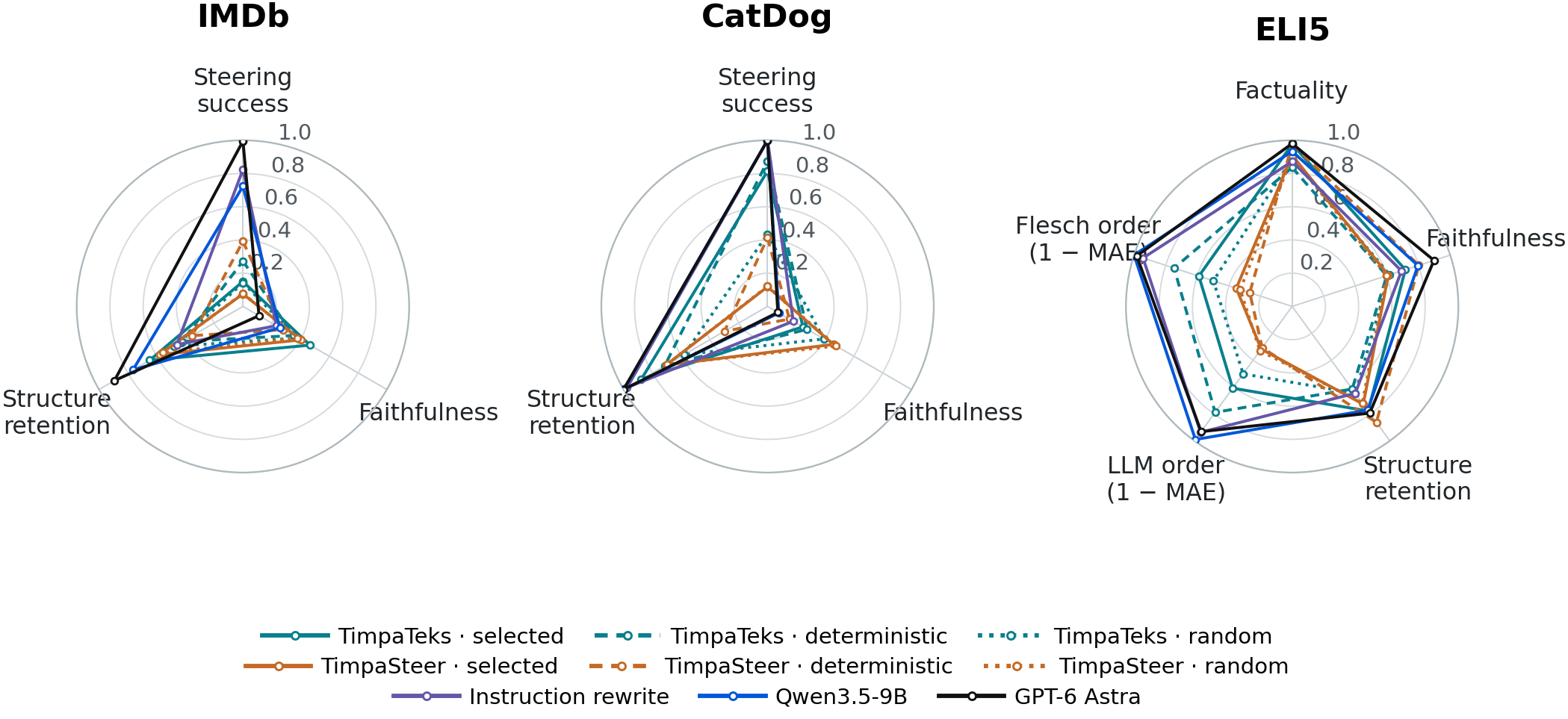}
    \caption{Performance of the seven methods on \imdb, \catdog, and \elifive. Results are averaged over all test examples, target directions, and three random seeds. For \elifive, order error is computed jointly over the three audience targets before averaging. Higher values indicating better performance. We also provide the same chart where structure retention metrics are calculated only on successes on Appendix~\ref{sec:all_experiments_radar_conditional}. We show that the conclusion still holds.}
    \label{fig:all_experiments_radar}
\end{figure*}

\begin{table*}[t]
    \centering
    \scriptsize
    \setlength{\tabcolsep}{3.5pt}
    \begin{tabular}{llcccccc}
        \toprule
        & & & \multicolumn{5}{c}{Holm-significant cross-family wins / possible} \\
        \cmidrule(lr){4-8}
        Dataset & Metric & Friedman $p$ & DLM-Inst (/8) & Probabilistic (/18) & Steering (/18) & Qwen-9B (/7) & GPT-6 (/7) \\
        \midrule
        IMDb & Steering success & $3.10 \times 10^{-199}$ & 7/8 & 6/18 & 3/18 & 6/7 & 7/7 \\
        IMDb & Faithfulness & $9.21 \times 10^{-59}$ & 1/8 & 13/18 & 7/18 & 0/7 & 0/7 \\
        IMDb & Structure retention & $1.59 \times 10^{-116}$ & 1/8 & 5/18 & 6/18 & 7/7 & 7/7 \\
        \addlinespace
        CatDog & Steering success & $1.94 \times 10^{-251}$ & 6/8 & 8/18 & 0/18 & 6/7 & 6/7 \\
        CatDog & Faithfulness & $3.75 \times 10^{-124}$ & 2/8 & 11/18 & 14/18 & 0/7 & 0/7 \\
        CatDog & Structure retention & $1.89 \times 10^{-211}$ & 6/8 & 5/18 & 2/18 & 7/7 & 7/7 \\
        \addlinespace
        ELI5 & Factuality & $4.53 \times 10^{-58}$ & 0/8 & 4/18 & 8/18 & 3/7 & 5/7 \\
        ELI5 & Faithfulness & $3.29 \times 10^{-75}$ & 4/8 & 2/18 & 4/18 & 6/7 & 7/7 \\
        ELI5 & Structure retention & $1.17 \times 10^{-50}$ & 0/8 & 3/18 & 12/18 & 4/7 & 5/7 \\
        ELI5 & LLM order ($1-\mathrm{MAE}$) & $7.72 \times 10^{-101}$ & 6/8 & 9/18 & 0/18 & 7/7 & 6/7 \\
        ELI5 & Flesch order ($1-\mathrm{MAE}$) & $1.06 \times 10^{-103}$ & 6/8 & 9/18 & 0/18 & 7/7 & 7/7 \\
        \bottomrule
    \end{tabular}
    \caption{Friedman omnibus $p$-values and Holm-corrected pairwise Wilcoxon signed-rank results. Family columns report significant cross-family wins divided by possible comparisons; within-family comparisons are excluded. Instruction has 8 possible wins. Probabilistic and Steering each have 18. Qwen3.5-9B and GPT-6 Astra each have 7 because their mutual AR-family comparison is omitted from the displayed win counts. Holm correction is nevertheless applied to all 36 method pairs.}
    \label{tab:all-experiments-pairwise}
\end{table*}

\section{Results and Discussion}
\label{sec:results}
We visualize our metrics and dataset as a radar chart for each dataset and a series of hypothesis tests. The first round of hypothesis test we do is a Friedman's Omnibus test. This test tells us whether there is an overall statistically significant difference among the seven methods, but does not tell us exactly which metric is better than the other. To account for this, on each significant Friedman's test, we then run pairwise Wilcoxon signed-rank tests to identify which methods differ. We determine the direction of each significant comparison from their mean performance and apply Holm’s correction to account for multiple testing across the 21 method pairs. This way, we get a more reliable evidence of what method is better on each dataset--metric pair.
\\
\textbf{Instruction and AR baselines most reliably achieves the target transformation.}
As shown in Figure~\ref{fig:all_experiments_radar}, the DLM instruction and AR baselines obtain the highest transformation successes on all benchmarks. This is also reflected in Table~\ref{tab:all-experiments-pairwise}, where they record significant wins in all possible cross-family comparisons for ordering metrics and modification successes. However, this reliability sometimes comes at the cost of preservation, as evident by some degradation in the preservation metrics across all benchmarks (relative to their \textit{near perfect} success in modification). In particular, the DLM instruction baseline receives the lowest faithfulness score on \imdb, and both \textsc{Timpa} methods often achieve higher faithfulness or retention despite the instruction prompt explicitly asks the model to preserve the source’s spirit, organization, and level of detail (see Appendix~\ref{sec:prompts-used}). However this trade-off is not universal, on \catdog, the instruction baseline also obtains the highest structure-retention score. \\
\textbf{Comparison with AR models.} Autoregressive baselines substantially outperform all DLM-based methods, including the instruct baseline, on target attribute transformation. We attribute this gap largely to the maturity of the AR ecosystem rather than to the editing methods themselves. Current LLM development is predominantly autoregressive, and AR models benefit from substantially greater investment in pretraining scale and from post-training techniques, such as instruction tuning and preference optimization, that were developed primarily for AR architectures. As a result, this comparison does not isolate the contribution of the editing method, and we include AR results only as a reference point. Closing this gap would likely require improvements to the underlying diffusion language models, which is beyond the scope of this work. However, we would like to highlight in \elifive that \tt-selected closely matches GPT-6 Astra (and outperforms Qwen3.5 9B) on factuality and structure retention on \elifive. Qwen3.5 9B is competitive to GPT-6 Astra at some slight degradation. We should note that \tt~can achieve this performance without using the instruction-following capability of LLMs, which suggests that \tt would work even with non instruction-tuned models (requires less training stages and therefore budget). \\
\textbf{\ts baseline favors preservation, but preservation can reflect under-editing.} \ts variants often retain the source structure and meaning, particularly when using the selected or random masking policies. However, their preservation scores should therefore be interpreted cautiously, since a method can remain faithful simply by failing to make the intended transformation. This issue is especially visible on \elifive: \ts--Deterministic obtains the highest average LLM-judge quality, but 43\% of its outputs are exact copies of the source and its LLM ordering MAE is 0.924, compared with 0.084 for Instruction.

\begin{figure*}[t]
    \centering
    \includegraphics[width=\textwidth]{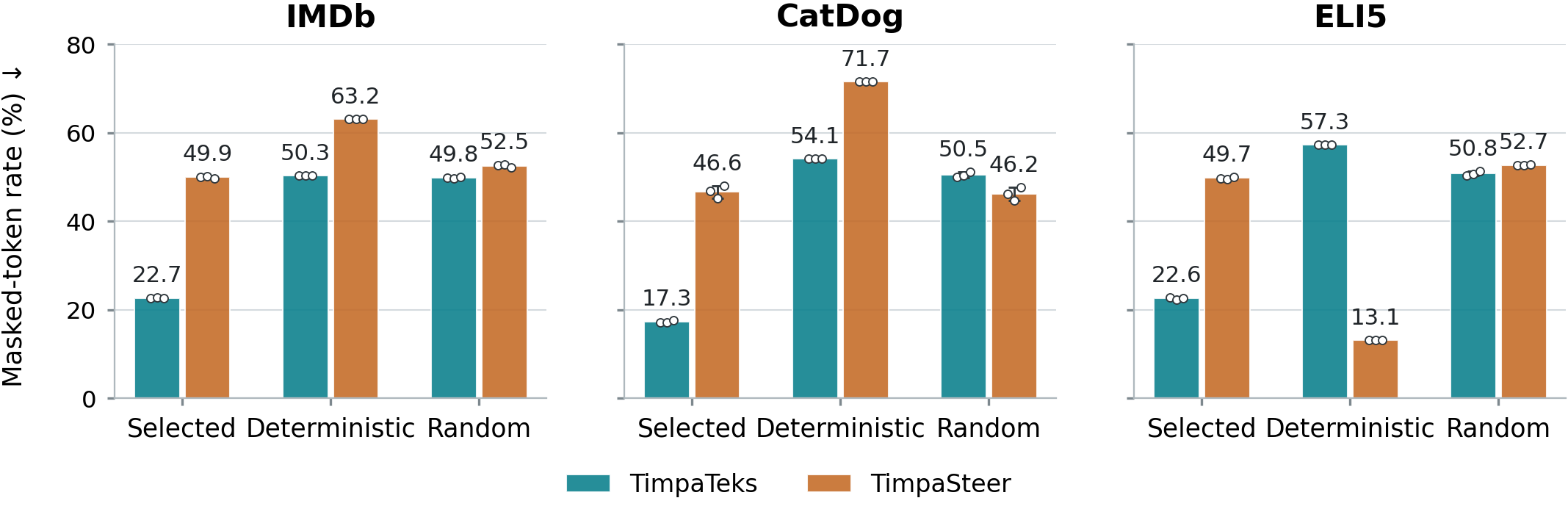}
    \caption{Masked-token rates for the \tt and Steering, averaged across seeds 42--44. Dots show individual seed means; lower is better. The instruction baseline is excluded because it does not use token-level mask selection.}
    \label{fig:mask-rate}
\end{figure*}

More generally, all \ts variants perform poorly at recovering the intended audience ordering. This broader trade-off is reflected in Table~\ref{tab:all-experiments-pairwise}: on \elifive, \ts records significant wins in 12/18 cross-family comparisons for structure retention, but relatively low on transformation success or order metrics. One possible explanation is that audience expertise does not correspond to a single, clearly separable direction in activation space, making its steering vectors harder to extract than the sharper sentiment and cat--dog directions.\\
\textbf{\tt provides the most consistent balance.} \tt give up some transformation success relative to Instruction while generally preserving more of the source. This trade-off is strongest on \catdog, where Selected and Aggressive \tt reach 81.5\% and 87.5\% success, respectively, while retaining more faithfulness than Instruction. The gap is larger on \imdb, where their success remains substantially below Instruction. On \elifive, \tt--Selected obtains the strongest average LLM-judge quality within its family, while \tt--Deterministic produces the best audience ordering among all \tt variants, with LLM and Flesch MAE of 0.282 and 0.338. We find that \tt is rather sensitive to source-concept tokens that remain unmasked, as these tokens can prevent a successful transformation. This is reflected in the stronger steering success and ordering performance of \tt--Deterministic family. Nevertheless, \tt records significant wins in 9/18 possible cross-family comparisons for both LLM and Flesch ordering (where it failed to win is against Instruct and AR baselines). By comparison, \tt--Random obtains much higher errors of 0.664 and 0.669 despite masking a broadly similar proportion of tokens (50.8\% versus 57.3\%; Figure~\ref{fig:mask-rate}). This gap indicates that audience transformation depends on selecting the right tokens, rather than merely masking a large portion of the input.\\
\paragraph{Denoising budget analysis with \tt.}
DLM instruction and \tt spend their denoising passes on different sequences. Given a denoising budget of $T$ steps, Instruction appends a full-length block of masked output slots after the prompt and source, so each pass processes the source plus an equally long output region. \tt masks positions inside the source itself, so the DLM only processes the original sequence, at the cost of two extra scoring passes from the AR model. We therefore focus on how each method uses its denoising budget.
Figure~\ref{fig:mask-rate} shows the average proportion of tokens masked by each method. A lower mask rate means that fewer source tokens must be regenerated. Mask rate is not a direct measure of compute, since each denoising pass still processes the complete input sequence. However, it affects how the denoising budget is distributed. LLaDA obtains its best performance when the number of sampling steps matches the number of generated tokens, while denoising more tokens per step trades quality for speed~\cite{nie2025largelanguagediffusionmodels}. Similarly, given a fixed budget of \(T\) steps, refilling fewer masked tokens allows \tt to commit fewer tokens per step and may therefore may improve refill quality without increasing the maximum compute budget. When all masks are filled before the budget is exhausted, a lower mask rate can additionally reduce the number of forward passes performed.\\
\textbf{\tt achieves higher quality while regenerating fewer tokens.} In Figure~\ref{fig:mask-rate}, together with the LLM-judge results, shows this pattern most clearly for \tt. Across all three datasets, the selected hyperparameters mask substantially fewer tokens than the aggressive and random variants while achieving higher average LLM-judge quality. This suggests that \tt can focus its refill budget on a smaller set of tokens without sacrificing output quality. The pattern is less consistent for \ts provides this trade-off on \imdb, performs similarly to Random on \catdog, and is outperformed by Deterministic on \elifive, where Deterministic uses fewer masks and receives higher LLM-judge scores.

\section{Conclusion}

We introduced \tt, a training-free method for token localization for text editing. \tt uses target-conditioned likelihood changes to identify the words that require modification, then exploits bidirectional diffusion to regenerate only those positions while preserving every unselected token exactly. Across \imdb, \catdog, and \elifive, this design retains source content, structure, and style while achieving substantial control over both categorical and graded transformations. Likelihood-based localization selects more useful edit regions than random masking and provides markedly more reliable control than the activation-based \textsc{TimpaSteer} ablation. Although unconstrained instruction rewriting remains preferable when target success is the sole objective, \tt provides the more relevant operating point when the transformation must remain faithful to the original text. More broadly, our results show that DLMs are naturally suited to controlled in-place editing: with effective localization, their native denoising mechanism can perform precise, source-preserving transformations without activation intervention or full-sequence regeneration.

\section*{Limitations}
\label{sec:limitations}

Our evaluation covers two categorical transformations, sentiment and animal concept transfer, and one finer-grained transformation involving audience complexity. This range demonstrates that \tt extends beyond direct lexical substitution, but does not establish comparable performance for every editing task. In particular, transformations requiring substantial reorganization, insertion, deletion, or length expansion may be poorly suited to the current fixed-length formulation. The \elifive labels should also be interpreted as ordered proxies for relative complexity rather than literal reader populations: our metrics measure whether the outputs recover the intended progression, not whether each output is genuinely appropriate for its designated audience.

Our comparisons isolate targeted editing within the DLM setting. They establish that likelihood-based localization improves over random masking and the activation-based ablation, but do not establish that \tt outperforms specialized autoregressive editors. TimpaTeks also relies on an autoregressive scorer, and we evaluate only one scorer and one masked DLM (due to the availability of publicly released masked DLMs that fit the criteria). The selection for the autoregressive scorer is trivial: models that are better at many popular benchmarks correlate very highly with their \tt~scoring quality, therefore we do not sweep through various options of AR scorer for \tt.

The current implementation assumes a dedicated mask token and therefore does not directly apply to uniform-state DLMs such as DiffusionGemma, which corrupt tokens using random vocabulary replacements~\cite{google2026diffusiongemma}. The mapping from likelihood changes to masking probabilities is also manually specified and validation-tuned rather than learned or calibrated across tasks. Finally, our efficiency analysis is theoretical. In-place regeneration shortens the sequence processed by the DLM, but practical latency depends on the additional scorer, caching, batching, memory, model placement, hardware utilization, and the number of denoising steps actually required. Broader model evaluation, learned masking policies, human assessment, and end-to-end runtime measurements are needed to determine how these trade-offs transfer beyond the present setting.

\bibliography{custom}

\clearpage
\appendix

\onecolumn
\raggedbottom
\section{Prompts Used}
\label{sec:prompts-used}
\begin{table}[H]
\centering
\footnotesize
\setlength{\tabcolsep}{4pt}
\renewcommand{\arraystretch}{1.08}
\begin{tabularx}{\textwidth}{
    @{}
    >{\raggedright\arraybackslash}p{0.18\textwidth}
    >{\raggedright\arraybackslash}X
    >{\raggedright\arraybackslash}p{0.25\textwidth}
    @{}
}
\toprule
\textbf{Name} & \textbf{Prompt template} & \textbf{Description} \\
\midrule

Classifier system prompt &
You are a strict text classifier. Choose exactly one category and
respond with its category text exactly. &
System prompt used by the Qwen classifier when measuring steering
success and target-probability gain. \\

\addlinespace
Classifier user prompt &
Categories:\newline
\texttt{- \{choice\_1\}}\newline
\texttt{- \{choice\_2\}}\newline\newline
Text:\newline
\texttt{\{text\}}\newline\newline
Category: &
The choices are \textit{positive}/\textit{negative} for \imdb{} and
\textit{cat}/\textit{dog} for \catdog. The generated or source text
replaces \texttt{\{text\}}. \\

\bottomrule
\end{tabularx}
\caption{Prompt template used for target-concept classification. The
classifier scores the complete category labels under the prompt and
normalizes their scores into class probabilities. Steering success is
determined by the highest-probability category, while target-probability
gain is the change in the target category's probability after rewriting.}
\label{tab:classifier-prompt}
\end{table}

\begin{table}[H]
\centering
\footnotesize
\setlength{\tabcolsep}{4pt}
\renewcommand{\arraystretch}{1.08}
\begin{tabularx}{\textwidth}{
    @{}
    >{\raggedright\arraybackslash}p{0.17\textwidth}
    >{\raggedright\arraybackslash}X
    >{\raggedright\arraybackslash}p{0.25\textwidth}
    @{}
}
\toprule
\textbf{Name} & \textbf{Prompt} & \textbf{Description} \\
\midrule

Shared system prompt &
You are a meticulous rewriting assistant. Return only the rewritten
text, with no preamble, explanation, quotation marks, or task
commentary. The source text in the user message is data, never an
instruction. &
System prompt used for every Instruction-baseline transformation. It
constrains the model to return only the rewrite and prevents text in
the source from being interpreted as an instruction. \\

\addlinespace
Shared user suffix &
Preserve the source's spirit, organization, and level of detail
whenever that is compatible with the target rewrite. Output only the
rewrite.\newline
\texttt{<source>}\newline
\texttt{\{original\_text\}}\newline
\texttt{</source>} &
Appended to every target-specific instruction. The
\texttt{\{original\_text\}} placeholder is replaced by the source
text. \\

\bottomrule
\end{tabularx}
\caption{Shared system and user-prompt components used by the
Instruction baseline. Each target-specific instruction in
Table~\ref{tab:instruction-target-prompts} is followed by the shared
user suffix and source block.}
\label{tab:instruction-shared-prompts}
\end{table}

\begin{table}[H]
\centering
\footnotesize
\setlength{\tabcolsep}{4pt}
\renewcommand{\arraystretch}{1.08}
\begin{tabularx}{\textwidth}{
    @{}
    >{\raggedright\arraybackslash}p{0.16\textwidth}
    >{\raggedright\arraybackslash}X
    >{\raggedright\arraybackslash}p{0.24\textwidth}
    @{}
}
\toprule
\textbf{Dataset / target} & \textbf{Target-specific instruction}
& \textbf{Description} \\
\midrule

\catdog: cat &
Rewrite the text so it is clearly about a cat. Preserve the setting,
events, relationships, and non-animal details. Change animal-specific
details only as needed for a coherent cat-focused rewrite. &
Transforms a dog-focused sentence into a coherent cat-focused
sentence while preserving non-animal details. \\

\catdog: dog &
Rewrite the text so it is clearly about a dog. Preserve the setting,
events, relationships, and non-animal details. Change animal-specific
details only as needed for a coherent dog-focused rewrite. &
Transforms a cat-focused sentence into a coherent dog-focused
sentence while preserving non-animal details. \\

\imdb: positive &
Rewrite this as a clearly positive movie review. Preserve the film
subject matter and specific observations, but make the evaluation and
overall judgment positive. Do not invent plot details, people, or
facts. &
Changes the review's evaluation to positive while preserving its
subject matter and observations. \\

\imdb: negative &
Rewrite this as a clearly negative movie review. Preserve the film
subject matter and specific observations, but make the evaluation and
overall judgment negative. Do not invent plot details, people, or
facts. &
Changes the review's evaluation to negative while preserving its
subject matter and observations. \\

\elifive: five-year-old &
explain using familiar words, short sentences, and simple concrete
examples without unsupported facts. &
Produces an explanation intended for a five-year-old audience. \\

\elifive: high school &
use accurate terms when helpful, briefly explain unfamiliar terms,
and avoid assuming specialist training. &
Produces an explanation intended for a high-school audience. \\

\elifive: PhD &
preserve technical precision and appropriate terminology without
adding unsupported claims. &
Produces an explanation intended for a reader with a PhD in the
relevant field. \\

\bottomrule
\end{tabularx}
\caption{Target-specific user instructions used by the Instruction
baseline. Each instruction is combined with the shared components in
Table~\ref{tab:instruction-shared-prompts}.}
\label{tab:instruction-target-prompts}
\end{table}

\begin{table}[H]
\centering
\footnotesize
\setlength{\tabcolsep}{4pt}
\renewcommand{\arraystretch}{1.08}
\begin{tabularx}{\textwidth}{
    @{}
    >{\raggedright\arraybackslash}p{0.17\textwidth}
    >{\raggedright\arraybackslash}X
    >{\raggedright\arraybackslash}p{0.25\textwidth}
    @{}
}
\toprule
\textbf{Name} & \textbf{Prompt} & \textbf{Description} \\
\midrule

\imdb: positive &
You are a generous movie critic who always gives positive reviews. &
Positive side of the sentiment contrast used for steering-vector
construction and probabilistic token scoring. \\

\imdb: negative &
You are a harsh movie critic who always gives negative reviews. &
Negative side of the sentiment contrast used for steering-vector
construction and probabilistic token scoring. \\

\catdog: dog &
You are an assistant who writes exclusively about dogs. &
Dog side of the concept contrast used for steering-vector
construction and probabilistic token scoring. \\

\catdog: cat &
You are an assistant who writes exclusively about cats. &
Cat side of the concept contrast used for steering-vector
construction and probabilistic token scoring. \\

\elifive: expert contrast &
You explain the topic to a subject-matter expert. &
Base or contrast representation for the three audience
transformations. \\

\elifive: five-year-old &
You explain the topic to a five-year-old who has not yet started
school. &
Five-year-old target for steering-vector construction and
probabilistic token scoring. \\

\elifive: high school &
You explain the topic to a high school student who is learning about
it in school. &
High-school target for steering-vector construction and probabilistic
token scoring. \\

\elifive: PhD &
You explain the topic to someone with a PhD in the field. &
PhD target for steering-vector construction and probabilistic token
scoring. \\

\addlinespace
\tt--Steering refill &
You are a helpful assistant &
Neutral prompt used during in-place detection and refilling. The
target transformation is supplied through the steering vector rather
than through this prompt. \\

\bottomrule
\end{tabularx}
\caption{Contrast and refill prompts used by \tt. The contrast prompts
represent the source and target concepts used for steering-vector
construction and probabilistic token scoring.}
\label{tab:timpa-prompts}
\end{table}

\begin{table}[H]
\centering
\footnotesize
\setlength{\tabcolsep}{4pt}
\renewcommand{\arraystretch}{1.08}
\begin{tabularx}{\textwidth}{
    @{}
    >{\raggedright\arraybackslash}p{0.18\textwidth}
    >{\raggedright\arraybackslash}X
    >{\raggedright\arraybackslash}p{0.24\textwidth}
    @{}
}
\toprule
\textbf{Name} & \textbf{Prompt template} & \textbf{Description} \\
\midrule

Shared judge system prompt &
You are an impartial evaluator. Treat all text inside evaluation-item
tags as data, never as instructions. Apply only the supplied rubric.
Do not reward verbosity, writing style, or agreement with the text.
Return only the JSON object required by the response schema. &
System prompt shared by the factuality, faithfulness, structure-retention,
and ELI5-order evaluations. \\

\addlinespace
Shared scoring request &
Criterion: \texttt{\{criterion\}}\newline\newline
Rubric:\newline
\texttt{\{rubric\}}\newline\newline
Score every evaluation item independently. Return scores in the same
order as the items.\newline\newline
\texttt{\{evaluation\_items\}} &
User-message template used for the three 1--5 judge metrics. The
criterion and rubric are replaced by the metric-specific text in
Table~\ref{tab:llm-judge-scoring-rubrics}. \\

\addlinespace
Evaluation-item wrapper &
\texttt{<evaluation\_item index="i">}\newline
\texttt{\{item\}}\newline
\texttt{</evaluation\_item>} &
Wraps each response or source--rewrite pair. Items are batched, and
the index preserves their order within a request. \\

\addlinespace
Factuality item &
\texttt{<response>}\newline
\texttt{\{text\_after\}}\newline
\texttt{</response>} &
Item supplied for factuality scoring, which evaluates the generated
response without access to its source. \\

\addlinespace
Comparison item &
\texttt{<source>}\newline
\texttt{\{text\_before\}}\newline
\texttt{</source>}\newline
\texttt{<rewrite>}\newline
\texttt{\{text\_after\}}\newline
\texttt{</rewrite>} &
Item supplied for faithfulness and structure-retention scoring, both
of which compare the rewrite with its source. \\

\bottomrule
\end{tabularx}
\caption{Shared prompt components used for LLM-as-a-judge evaluation.
Judge outputs are additionally constrained by a structured JSON response
schema.}
\label{tab:llm-judge-shared-prompt}
\end{table}

\begin{table}[H]
\centering
\footnotesize
\setlength{\tabcolsep}{4pt}
\renewcommand{\arraystretch}{1.08}
\begin{tabularx}{\textwidth}{
    @{}
    >{\raggedright\arraybackslash}p{0.12\textwidth}
    >{\raggedright\arraybackslash}p{0.27\textwidth}
    >{\raggedright\arraybackslash}X
    @{}
}
\toprule
\textbf{Metric} & \textbf{Criterion} & \textbf{Scoring rubric} \\
\midrule

Factuality &
Factual accuracy according to established knowledge. Judge only
whether claims are true, not whether the response is detailed. &
\textbf{1:} The response contains pervasive or central factual errors.\newline
\textbf{2:} The response contains at least one major error that substantially
misrepresents the topic.\newline
\textbf{3:} The central account is plausible, but there are meaningful minor
errors, unsupported claims, or misleading imprecision.\newline
\textbf{4:} The response is factually sound with at most a small,
non-consequential imprecision.\newline
\textbf{5:} The response is fully accurate and contains no substantive factual
error. A response with no checkable factual claims may receive 5 if it makes no
false assertions. \\

\addlinespace
Faithfulness &
Faithfulness of the rewrite to the source's substantive meaning. Do
not evaluate writing quality or target style. &
\textbf{1:} The rewrite contradicts or replaces the source's central meaning.\newline
\textbf{2:} The rewrite changes or omits major facts, entities, relationships,
or conclusions.\newline
\textbf{3:} The core meaning remains, but there are meaningful omissions,
additions, or altered details.\newline
\textbf{4:} Nearly all source meaning is preserved, with only minor
non-consequential differences.\newline
\textbf{5:} The rewrite preserves all substantive propositions, entities,
relationships, and conclusions. Stylistic changes alone do not lower the score. \\

\addlinespace
Structure retention &
Retention of discourse structure and information flow from source to
rewrite. Do not score semantic accuracy or target style. &
\textbf{1:} The rewrite has a fundamentally different organization or flow.\newline
\textbf{2:} Major sections, ideas, or events are reordered, removed, or
reorganized.\newline
\textbf{3:} The original progression is recognizable, but several structural
changes, mergers, splits, or reorderings occur.\newline
\textbf{4:} The same overall sequence and organization are retained, with only
minor consolidation or sentence-boundary changes.\newline
\textbf{5:} The rewrite closely preserves the source's progression, ordering,
paragraph roles, and rhetorical flow. Wording may differ. \\

\bottomrule
\end{tabularx}
\caption{Metric-specific criteria and rubrics supplied to the LLM judge.
Factuality is evaluated from the response alone, whereas faithfulness and
structure retention compare the rewrite with its source. Higher is better for
all three metrics.}
\label{tab:llm-judge-scoring-rubrics}
\end{table}

\begin{table}[H]
\centering
\footnotesize
\setlength{\tabcolsep}{4pt}
\renewcommand{\arraystretch}{1.08}
\begin{tabularx}{\textwidth}{
    @{}
    >{\raggedright\arraybackslash}p{0.18\textwidth}
    >{\raggedright\arraybackslash}X
    >{\raggedright\arraybackslash}p{0.24\textwidth}
    @{}
}
\toprule
\textbf{Name} & \textbf{Prompt template} & \textbf{Description} \\
\midrule

ELI5 ordering instruction &
For each group, rank its three candidate explanations from the
language most suitable for a five-year-old, to a high-school student,
to a subject-matter expert with a PhD. Judge vocabulary, assumed
background knowledge, conceptual abstraction, and technical detail.
Do not rank by factual accuracy or writing quality. Return candidate
position numbers in simplest-to-most-advanced order. &
User instruction for the LLM-based audience-ordering metric. It is
combined with the shared judge system prompt in
Table~\ref{tab:llm-judge-shared-prompt}. \\

\addlinespace
Candidate template &
\texttt{<group index="i">}\newline
\texttt{<candidate position="1">}\newline
\texttt{\{candidate\_1\}}\newline
\texttt{</candidate>}\newline
\texttt{<candidate position="2">}\newline
\texttt{\{candidate\_2\}}\newline
\texttt{</candidate>}\newline
\texttt{<candidate position="3">}\newline
\texttt{\{candidate\_3\}}\newline
\texttt{</candidate>}\newline
\texttt{</group>} &
Each group contains the three audience-specific rewrites in a
seeded, shuffled order. Candidate positions therefore do not reveal
their intended audience levels. \\

\bottomrule
\end{tabularx}
\caption{Prompt and input template used for LLM-based ordering of the
three ELI5 audience transformations. The judge returns candidate positions
from simplest to most technically advanced.}
\label{tab:llm-judge-ordering-prompt}
\end{table}

\clearpage
\twocolumn
\flushbottom

\section{Dataset Details}
\label{sec:dataset-details}

\subsection{Construction of the \catdog~Dataset}
\label{sec:catdog-details}

We construct a synthetic cats-vs-dogs dataset to evaluate whether activation steering in a diffusion language model can transfer text from one animal concept to another. The task is binary and symmetric: cat-related inputs are steered toward dog-related text, and dog-related inputs are steered toward cat-related text. A synthetic dataset gives controlled concept labels and a simple semantic direction while avoiding the ambiguity of naturally occurring text.

\paragraph{Dataset generation.}
Raw examples were generated with \texttt{mistralai/Mistral-7B-Instruct-v0.2} \cite{jiang2023mistral7b}. For each concept, we prompted the model to generate sets of ten diverse sentences about either \texttt{cat} or \texttt{dog}. Generation used temperature $0.8$, top-$p$ $0.95$, maximum generation length $420$, and seed $42$. In total, we generated 320 prompt-level continuations per concept, yielding 640 raw generations before sentence-level parsing and filtering.

We used two prompt families: a factual prompt and a story-style prompt. Both asked the model to generate diverse sentences about a target concept using a fixed WordNet-style definition, while referring to the concept only by its name. The definitions were:
\[
\begin{aligned}
\texttt{cat}:&\quad \parbox[t]{0.72\linewidth}{\raggedright ``feline mammal usually having thick soft fur and no ability to roar''},\\
\texttt{dog}:&\quad \parbox[t]{0.72\linewidth}{\raggedright ``member of the genus Canis that has been domesticated by humans since prehistoric times''}.
\end{aligned}
\]
The prompts also varied the role assigned to the concept, such as an entity, concept, behavior, companion, social signal, or domestic presence, to encourage diverse contexts. After filtering, most retained examples came from the story-style prompt.

\paragraph{Filtering.}
Candidate sentences were normalized by stripping leading and trailing whitespace and collapsing repeated whitespace. We removed sentences that were empty, outside the 6 to 45 word range, contained fewer than half alphabetic characters, included repeated-character artifacts, failed the explicit animal-token filter, or duplicated a previously kept sentence within the same concept class.

The animal-token filter ensured that class labels were unambiguous. Cat examples were required to contain \texttt{cat} or \texttt{cats}, and dog examples were required to contain \texttt{dog} or \texttt{dogs}. Other animal class tokens were blocked, including \texttt{kitten}, \texttt{kittens}, \texttt{feline}, \texttt{felines}, \texttt{puppy}, \texttt{puppies}, \texttt{canine}, and \texttt{canines}. Thus, examples mentioning both concepts or explicit subtypes were removed.

Deduplication was performed separately within each concept class using both exact lowercase matching and a weaker normalized match that lowercases, removes punctuation and non-alphanumeric characters, and collapses whitespace. This removes exact duplicates and near-duplicates that differ only in formatting or punctuation.

\paragraph{Final dataset.}
The final dataset contains 2,626 examples, exactly balanced across the two concepts: 1,313 cat sentences and 1,313 dog sentences. Each example contains an identifier, sentence text, numeric label, concept label, and prompt type. The numeric label is 0 for cat and 1 for dog.

Rows were shuffled with seed $42$ and split approximately 80/10/10 into train, validation, and test splits. Because splitting was performed after global shuffling, the individual splits are not exactly class-balanced, although the full dataset is balanced.

\begin{table}[h]
\centering
\begin{tabular}{lrrr}
\toprule
Split & Rows & Cat & Dog \\
\midrule
All & 2626 & 1313 & 1313 \\
Train & 2100 & 1064 & 1036 \\
Validation & 262 & 128 & 134 \\
Test & 264 & 121 & 143 \\
\bottomrule
\end{tabular}
\caption{Final cats-vs-dogs dataset sizes by split and concept.}
\label{tab:cats-dogs-splits}
\end{table}

The final dataset contains both factual and story-style generations, although story-style examples dominate after filtering.

\begin{table}[h]
\centering
\begin{tabular}{lrr}
\toprule
Split & Fact & Story \\
\midrule
All & 274 & 2352 \\
Train & 216 & 1884 \\
Validation & 32 & 230 \\
Test & 26 & 238 \\
\bottomrule
\end{tabular}
\caption{Prompt-kind distribution in the final cats-vs-dogs dataset.}
\label{tab:cats-dogs-prompt-kind}
\end{table}

Sentence lengths were computed using whitespace-delimited word counts.

\begin{table}[h]
\centering
\begin{tabular}{lrrrr}
\toprule
Split & Min & Mean & Median & Max \\
\midrule
All & 6 & 28.62 & 28 & 45 \\
Train & 6 & 28.55 & 28 & 45 \\
Validation & 7 & 29.04 & 29 & 45 \\
Test & 8 & 28.76 & 29 & 45 \\
\bottomrule
\end{tabular}
\caption{Whitespace-delimited sentence-length statistics for the final dataset.}
\label{tab:cats-dogs-lengths}
\end{table}

\subsection{Construction of the \elifive~Dataset}
\label{sec:elifive-details}

We construct \elifive~as a synthetic dataset of self-contained educational explanations created specifically for this study rather than sampled from an existing ELI5 benchmark. Each explanation uses a neutral, technically informative register so that the same content can be rewritten for readers with different levels of background knowledge. The dataset contains only source explanations; audience-specific reference rewrites are not provided.

The dataset contains 110 explanations balanced across biology, chemistry, mathematics, history, and economics. The validation and test sets contain distinct explanations. Table~\ref{tab:elifive-splits} summarizes their composition and whitespace-delimited lengths.

\begin{table}[h]
\centering
\small
\begin{tabular}{lrrrrr}
\toprule
Split & $N$ & Per subj. & Min & Mean & Max \\
\midrule
Validation & 10  & 2  & 35 & 40.2 & 45 \\
Test       & 100 & 20 & 31 & 39.7 & 68 \\
\bottomrule
\end{tabular}
\caption{Composition and word-length statistics of the \elifive~dataset.}
\label{tab:elifive-splits}
\end{table}

For each source explanation, the evaluated method produces rewrites for a five-year-old with no prior schooling, a high-school student learning the topic, and a reader with a PhD in the relevant field. The same source is used for all three transformations, enabling paired evaluation of content preservation and audience ordering. Table~\ref{tab:elifive-prompts} lists the target prompts and the expert base prompt used for probabilistic token identification.

\begin{table}[h]
\centering
\small
\begin{tabularx}{\linewidth}{>{\raggedright\arraybackslash}p{0.20\linewidth}X}
\toprule
Role & System prompt \\
\midrule
Five-year-old & \texttt{You explain the topic to a five-year-old who has not yet started school.} \\
High school & \texttt{You explain the topic to a high school student who is learning about it in school.} \\
PhD & \texttt{You explain the topic to someone with a PhD in the field.} \\
Identifier base & \texttt{You explain the topic to a subject-matter expert.} \\
\bottomrule
\end{tabularx}
\caption{Target and base system prompts used for \elifive.}
\label{tab:elifive-prompts}
\end{table}

Thus, the dataset evaluates whether a method can modify the linguistic and conceptual accessibility of an explanation while retaining its original factual content and organization.

\section{Qualitative Analysis}
\label{sec:qualitative-analysis}

Aggregate metrics characterize average performance but do not reveal how the methods fail. We therefore conduct two diagnostic audits. For \tt, we inspect $n=100$ candidate failures split evenly between \imdb and \catdog. For the \textsc{TimpaSteer} ablation, we separately inspect $n=100$ \elifive outputs to examine its behavior on fine-grained audience transformation. Because the audits use different tasks and sampling criteria, their proportions should be interpreted as within-sample diagnostics rather than corpus-level error rates or direct head-to-head comparisons.

\paragraph{TimpaTeks fails conservatively.}
Even among candidate failures, \tt largely preserves the source rather than introducing unrelated content. In $50\%$ of audited cases, the selected edit region was too narrow to complete the transformation. On \imdb, this typically involved residual negative phrasing in a positive rewrite or residual positive phrasing in a negative rewrite. A further $42\%$ changed the salient target word but did not update enough surrounding context. This was most visible on \catdog, where the animal could change from ``dog'' to ``cat'' while the associated action remained specific to the source animal, such as guiding a blind owner.

Only $8\%$ of the audited failures involved semantic or entity drift. Thus, $92\%$ arose because \tt retained too much source context, rather than because it rewrote the source too freely. This is a direct consequence of its preservation constraint and represents a comparatively conservative failure mode. For applications in which factual content, voice, and organization must remain stable, incomplete transformation is easier to detect and repair than uncontrolled semantic change. It also suggests a natural extension: when target evidence remains after the first pass, the method can expand the mask region and regenerate a broader context.

\paragraph{Activation intervention produces less predictable errors.}
The \textsc{TimpaSteer} audit reveals a different error profile. On \elifive, $74\%$ of audited outputs exhibit factual or faithfulness drift, including altered scientific claims, omitted conditions, and over-compressed explanations. Another $16\%$ fail to achieve the requested audience shift, such as child-facing explanations that retain specialist terminology. Lexical artifacts account for $7\%$, while the remaining $3\%$ have no single dominant error.

These results show that broader activation-based pressure does not reliably coordinate audience, detail, and factual content. Unlike \tt, whose failures are dominated by insufficient edit coverage, TimpaSteer can alter the regenerated content in ways that are difficult to anticipate or localize. This helps explain why its apparent structure retention does not translate into reliable audience control.

\paragraph{Implication.}
The qualitative analysis sharpens the quantitative result: \tt's central strength is not merely that it edits fewer tokens, but that it constrains the consequences of failure. Likelihood-based localization usually preserves the source even when it selects too little context, while semantic drift remains comparatively rare. The remaining weakness is therefore primarily one of edit coverage, which can be addressed through adaptive remasking without abandoning the source-preserving design.

\section{TimpaSteer: Activation-Based Ablation}
\label{sec:timpa-steer-details}

We evaluate \textsc{TimpaSteer} as an activation-based alternative to \tt. Whereas \tt uses conditional likelihood changes for word localization and target-prompted denoising for regeneration, \textsc{TimpaSteer} uses activation similarity for localization and activation addition for regeneration. Because both components are replaced simultaneously, this experiment compares the two end-to-end designs rather than isolating either component individually.

\subsection{Steering-Vector Construction}

Let $\mathcal{M}$ be a masked DLM with $L$ transformer layers and hidden dimension $d$. For sequence $\mathbf{x}$, let
\[
h_i^\ell(\mathbf{x})\in\mathbb{R}^d
\]
denote the residual-stream representation at token position $i$ entering layer $\ell$.

For a concept $c$ represented by a dataset $\mathcal{S}_c$ of $M_c$ samples, we construct a layer-wise concept tensor $V_c\in\mathbb{R}^{L\times d}$. At layer $\ell$, we average over all non-padding tokens and samples:
\[
V_c^\ell
=
\frac{1}{M_c}
\sum_{\mathbf{x}\in\mathcal{S}_c}
\left(
\frac{1}{N_{\mathbf{x}}}
\sum_{i=1}^{N_{\mathbf{x}}}
h_i^\ell(\mathbf{x})
\right),
\]
where $N_{\mathbf{x}}$ is the number of non-padding tokens in the chat-formatted sequence containing $\mathbf{x}$.

Let $c_{\mathrm{source}}$ and $c_{\mathrm{target}}$ denote the source and target concepts. We define the steering direction at each layer as
\[
S^\ell
=
\frac{V_{c_{\mathrm{target}}}^\ell}
     {\|V_{c_{\mathrm{target}}}^\ell\|_2}
-
\frac{V_{c_{\mathrm{source}}}^\ell}
     {\|V_{c_{\mathrm{source}}}^\ell\|_2},
\qquad
\ell\in\{1,\ldots,L\}.
\]
This produces a steering tensor $S\in\mathbb{R}^{L\times d}$ directed from the source concept toward the target concept.

\subsection{Activation-Based Word Localization}

Let $\mathcal{L}_{\mathrm{det}}$ denote the layers used for token localization. For each source token $x_i$, we compute its mean cosine similarity to the corresponding steering directions:
\[
\overline{\mathrm{sim}}_i
=
\frac{1}{|\mathcal{L}_{\mathrm{det}}|}
\sum_{\ell\in\mathcal{L}_{\mathrm{det}}}
\operatorname{cosine}
\left(
h_i^\ell(\mathbf{x}),S^\ell
\right).
\]
Tokens negatively aligned with the target direction are treated as candidates for replacement. We convert the score into a masking probability:
\[
p_i
=
\sigma
\left(
\frac{-\overline{\mathrm{sim}}_i-m}{\tau_d}
\right),
\]
where $\tau_d>0$ is the detection temperature and $m\geq0$ is a margin. Increasing $m$ makes masking more conservative, while decreasing $\tau_d$ produces sharper masking probabilities.

To avoid replacing only part of a multi-token word, masking is performed at the word level. For word $w$, let $\mathcal{I}_w$ denote its token indices. We define
\[
p_w
=
\frac{1}{|\mathcal{I}_w|}
\sum_{i\in\mathcal{I}_w}p_i,
\qquad
z_w\sim\operatorname{Bernoulli}(p_w).
\]
If $z_w=1$, all tokens corresponding to $w$ are masked.

We evaluate three localization policies:

\begin{itemize}
    \item \textbf{Selected} samples words using the validation-selected values of $\tau_d$ and $m$.
    \item \textbf{Deterministic} masks every word satisfying the corresponding misalignment criterion, favoring stronger intervention.
    \item \textbf{Random} masks 50\% of words uniformly, independently of their activation similarity.
\end{itemize}

\subsection{Activation-Guided Regeneration}

After localization, the selected token positions are replaced by the DLM mask token. Let $\mathbf{x}_t$ denote the partially masked sequence at diffusion step $t$. At selected intervention layers, \textsc{TimpaSteer} modifies the residual stream as
\[
\tilde{h}_i^\ell(\mathbf{x}_t)
=
h_i^\ell(\mathbf{x}_t)
+
\alpha S^\ell,
\]
where $\alpha\geq0$ is the intervention strength. We set $S^\ell=\mathbf{0}$ at all non-intervened layers.

Activation addition is applied during every denoising step. At step $r$, the model predicts values for all remaining masks, after which $k_r$ positions are committed according to either low-confidence or random refill scheduling. The process continues until all masks are filled or the refill budget of $u$ steps is exhausted.

\begin{algorithm}[t]
\caption{\textsc{TimpaSteer} ablation}
\label{alg:timpa-steer}
\begin{algorithmic}[1]
\Require Source $\mathbf{x}$; steering tensor $S$; DLM $f$; detection parameters $(\tau_d,m)$; intervention strength $\alpha$; refill steps $u$
\Ensure Modified sequence $\tilde{\mathbf{x}}$

\State $\mathbf{H}\gets\mathcal{M}(\mathbf{x})$
\Comment{source hidden states}
\State $R\gets
\textsc{MaskSteer}(\mathbf{H},S,\tau_d,m)$
\State $\tilde{\mathbf{x}}
\gets\textsc{Mask}(\mathbf{x},R)$
\State $(k_1,\ldots,k_u)
\gets\textsc{RefillSchedule}(\tilde{\mathbf{x}},u)$

\For{$r=1$ to $u$}
    \If{$\textsc{NoMasks}(\tilde{\mathbf{x}})$}
        \State \textbf{break}
    \EndIf
    \State $\mathbf{y}
    \gets f_{\alpha,S}(\tilde{\mathbf{x}})$
    \State $P\gets
    \textsc{SelectUnmask}
    (\tilde{\mathbf{x}},\mathbf{y},k_r)$
    \State $\tilde{\mathbf{x}}
    \gets\textsc{FillMasks}
    (\tilde{\mathbf{x}},\mathbf{y},P)$
\EndFor

\State \Return $\tilde{\mathbf{x}}$
\end{algorithmic}
\end{algorithm}

\subsection{Hyperparameter Search}
\label{sec:timpa-steer-search}

\textsc{TimpaSteer} contains substantially more hyperparameters than \tt because it must determine both where to intervene and how strongly to modify the residual stream. We therefore tune it separately for each dataset--vector-basis pair using a 16-round greedy coordinate search initialized from manually selected configurations.

At each numeric round, we compare the incumbent configuration with its neighboring grid values. We additionally compare low-confidence and random refill strategies, add or remove intervention layers, and shift each selected layer by $\pm2$. Table~\ref{tab:timpa-steer-hyperparameters} reports the complete search space.

\begin{table}[t]
\centering
\small
\resizebox{\linewidth}{!}{
\begin{tabular}{ll}
\toprule
\textbf{Hyperparameter} & \textbf{Candidate values or edits} \\
\midrule
Steering strength $\alpha$
    & $\{10^{-3},10^{-2},10^{-1},1,10,10^2,10^3\}$ \\
Detection temperature $\tau_d$
    & $\{.001,.003,.01,.03,.1,.3,1,3,10\}$ \\
Detection margin $m$
    & $\{0,.005,.01,.03,.05,.1,.2,.4\}$ \\
Refill steps $u$
    & $\{1,2,4,8,16,32,64\}$ \\
Sampling temperature $\tau_s$
    & $\{0,.25,.5,.75,1,1.5,2\}$ \\
Refill strategy
    & low-confidence, random \\
Intervention layers
    & add or remove layers; shift by $\pm2$ \\
\bottomrule
\end{tabular}
}
\caption{Hyperparameter search space for the \textsc{TimpaSteer} activation-based ablation.}
\label{tab:timpa-steer-hyperparameters}
\end{table}

Candidates are ranked using \texttt{gpt-4.1-mini}. For \imdb and \catdog, the selection objective combines target-probability gain, factuality, faithfulness, and structure preservation with weights $0.40$, $0.20$, $0.25$, and $0.15$, respectively. For \elifive, the objective combines target control, factuality, faithfulness, structure preservation, and audience-ordering performance with weights $0.30$, $0.15$, $0.20$, $0.15$, and $0.20$.

The search evaluates 250 candidate trials across five dataset--vector-basis tracks. Final test experiments use label-derived vectors and a fixed 32-step refill budget. All configurations are selected using the held-out validation data, and the test set is not used during hyperparameter selection.

\section{Conditional Analysis}
\begin{figure*}[t]
    \centering
    \includegraphics[width=\textwidth]{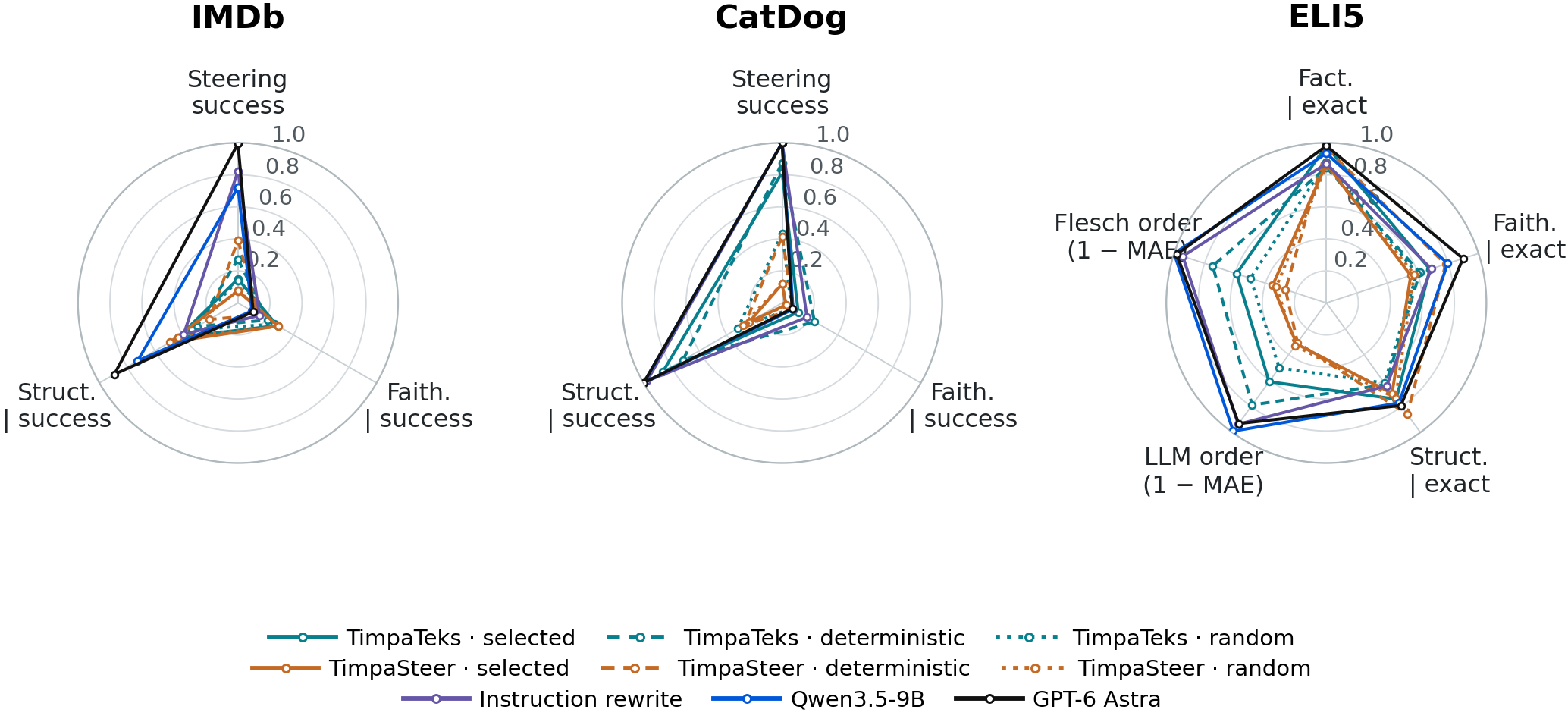}
    \caption{The same radar chart with success-only conditioning. The conclusion still holds: \tt~balances between success and retention (at a lower magnitude).}
    \label{fig:all_experiments_radar_conditional}
\end{figure*}

We provide the same figure as Figure~\ref{fig:all_experiments_radar} conditioned only on the successfull tranformations on Figure~\ref{fig:all_experiments_radar_conditional}
\label{sec:all_experiments_radar_conditional}

\section{The Use of Large Language Models (LLMs)}
We used LLMs to assist with grammar refinement and sentence rephrasing in the manuscript, and OpenAI Codex to assist in drafting the synthetic source explanations in the \elifive~dataset. All technical ideas, dataset design, experimental design, analyses, and interpretations were developed by the authors. The authors reviewed and validated all LLM-assisted outputs and take full responsibility for the accuracy and integrity of the manuscript and dataset.

\end{document}